\documentclass[letterpaper]{article} 
\usepackage[]{aaai2027}  
\usepackage{times}  
\usepackage{helvet}  
\usepackage{courier}  
\usepackage[hyphens]{url}  
\usepackage{graphicx} 
\usepackage{natbib}  
\usepackage{caption} 
\usepackage{amsmath,amsfonts,amssymb}
\usepackage{algorithmic}
\usepackage{algorithm}
\usepackage{array}
\usepackage{textcomp}
\usepackage{booktabs}
\usepackage{bm}
\usepackage{multirow}
\usepackage{makecell}
\usepackage{threeparttable}
\usepackage[table,xcdraw]{xcolor}
\newcommand{\method}{\mbox{4DR360$^\circ$}}

\newcommand{\detmethod}[1]{\makebox[5.0cm][c]{#1}}
\newcommand{\occmethod}[1]{\makebox[4.75cm][c]{#1}}
\newcommand{\modbox}[1]{\makebox[0.8cm][c]{#1}}
\newcommand{\detbox}[1]{\makebox[1.0cm][c]{#1}}
\newcommand{\detclsbox}[1]{\makebox[0.96cm][c]{#1}}
\newcommand{\occbox}[1]{\makebox[0.85cm][c]{#1}}
\newcommand{\mtoccbox}[1]{\makebox[0.95cm][c]{#1}}
\newcommand{\tabgap}{\noalign{\vskip 0.5pt}}

\newcommand{\cmark}{\checkmark}
\definecolor{best}{RGB}{255, 180, 180}     
\definecolor{second}{RGB}{255, 230, 180}   
\definecolor{third}{RGB}{255, 255, 200}    
\definecolor{lowbest}{RGB}{107, 174, 214}  
\definecolor{lowsecond}{RGB}{158, 202, 225}
\definecolor{lowthird}{RGB}{222, 235, 247}
\newcommand{\bestcell}[1]{\cellcolor{best}\textbf{#1}}
\newcommand{\secondcell}[1]{\cellcolor{second}\underline{#1}}
\newcommand{\thirdcell}[1]{\cellcolor{third}#1}

\definecolor{ncar}{RGB}{255, 165, 0}
\definecolor{npedestrian}{RGB}{128, 0, 128}
\definecolor{nrider}{RGB}{0, 0, 200}
\definecolor{nlarge_vehicle}{RGB}{220, 220, 0}
\definecolor{ntrailer}{RGB}{230, 230, 230}
\definecolor{nego_trailer}{RGB}{0, 0, 150}
\definecolor{ncycle}{RGB}{230, 230, 230}
\definecolor{nroad_obstacle}{RGB}{255, 69, 0}
\definecolor{ntraffic_fence}{RGB}{0, 0, 150}
\definecolor{ndriveable_surface}{RGB}{135, 206, 235}
\definecolor{nsidewalk}{RGB}{200, 200, 200}
\definecolor{notherflat}{RGB}{180, 180, 180}
\definecolor{nterrain}{RGB}{152, 251, 152}
\definecolor{nvegetation}{RGB}{34, 139, 34}
\definecolor{nmanmade}{RGB}{230, 230, 250}
\definecolor{diagblue}{RGB}{56, 98, 160}
\definecolor{diaggreen}{RGB}{58, 128, 92}
\definecolor{diagorange}{RGB}{185, 111, 42}
\definecolor{diagred}{RGB}{176, 77, 77}
\definecolor{diaggray}{RGB}{92, 98, 110}
\newcolumntype{L}[1]{>{\raggedright\arraybackslash}m{#1}}
\newcolumntype{C}[1]{>{\centering\arraybackslash}m{#1}}

\title{4DR360$^{\circ}$: State Reasoning for Joint 3D Detection and Occupancy Prediction \\ in
4D Radar-Camera Full-Scene Perception}
\author{
    Xiaokai Bai\textsuperscript{\rm 1},
    Lianqing Zheng\textsuperscript{\rm 2},
    Runwei Guan\textsuperscript{\rm 3},
    Songkai Wang\textsuperscript{\rm 1},
    Siyuan Cao\textsuperscript{\rm 1},
    Hui-liang Shen\textsuperscript{\rm 1}
}
\affiliations{
    \textsuperscript{\rm 1}College of Information Science and Electronic Engineering, Zhejiang University.\\
    \textsuperscript{\rm 2}School of Automotive Studies, Tongji University.\\
    \textsuperscript{\rm 3}Thrust of Artificial Intelligence, Hong Kong University of Science and Technology.\\
    shawnnnkb@gmail.com
}

\begin{document}

\maketitle
\begin{abstract}
Reliable autonomous driving requires full-scene perception that couples
foreground objects with dense semantic layout. Recently, 4D millimeter-wave
radar has emerged as a robust and affordable sensor, yet its sparse returns make
radar-camera fusion necessary for comprehensive scene understanding. Existing
radar-camera methods mainly optimize detection, while dual-task systems usually
decode boxes and occupancy with limited interaction. To address this gap and
advance radar-based multi-task learning, we propose \method, a 4D radar-camera
framework for 360$^\circ$ full-scene perception, which models semantic
occupancy as a persistent scene state rather than a terminal output. \method{} follows a cross-modal state reasoning paradigm, where
the occupancy state is modeled and propagated through stages for coarse-to-fine
feature aggregation. Specifically, State-guided BEV Enhancement (SBE)
strengthens intra-frame BEV representation, while Doppler-guided Temporal
Fusion (DTF) preserves state evidence over longer temporal horizons. Beyond the model, we
further extend ManTruckScenes with satellite-map-based generated occupancy
labels and pair it with OmniHD-Scenes in a unified cross-dataset
detection-and-occupancy protocol. The resulting experiments cover accuracy,
robustness, ablation, and efficiency under one radar-camera multi-task
evaluation framework. Code and labels will be released.
\end{abstract}

\section{Introduction}

Autonomous driving requires 360{$^\circ$} full-scene perception that jointly
models foreground agents and surrounding scene layout. In this setting, 3D
object detection estimates compact boxes for traffic participants
\cite{BEVFusion,SGDet3D,RaGS2026,R4Det2026}, whereas semantic occupancy
prediction recovers dense free, occupied, and semantic voxel states around the
ego vehicle \cite{MonoScene,VoxFormer,SurroundOcc,Occ3D2023,OmniHD}. These two
tasks describe the same scene at different granularities. Detection focuses on
instance-level localization, while occupancy supplies the spatial support and
layout context in which those instances exist. Therefore, reliable full-scene
perception requires a unified treatment of object reasoning and scene layout.

\begin{figure}[t]
    \centering
    \includegraphics[width=1.0\linewidth]{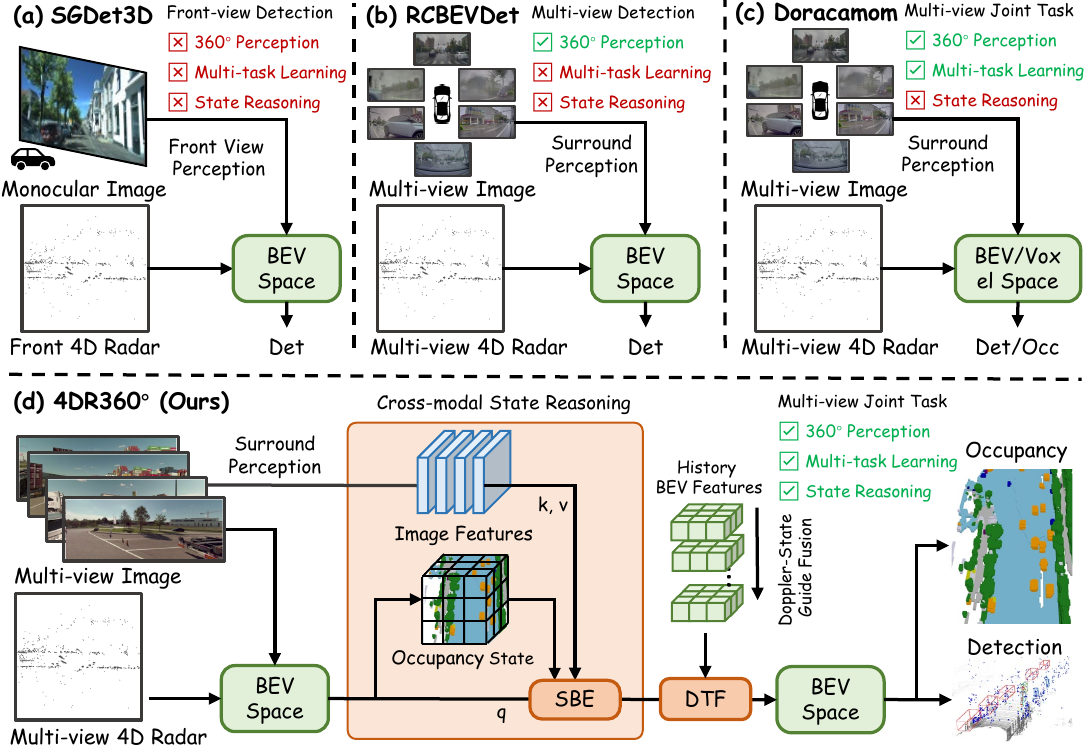}
    \caption{Comparison of representative 4D radar-camera paradigms. (a) SGDet3D
    is front-view and detection-only, (b) RCBEVDet extends detection to
    360$^\circ$ perception, (c) Doracamom supports joint detection and occupancy
    prediction without explicit state reasoning, and (d) \method{} organizes
    360$^\circ$ radar-camera multi-task perception through occupancy-state reasoning
    with SBE and DTF.}
    \label{fig_motivation}
    \vspace{-5pt}
\end{figure}
\begin{figure*}[t]
    \centering
    \includegraphics[pagebox=cropbox,width=1.0\linewidth]{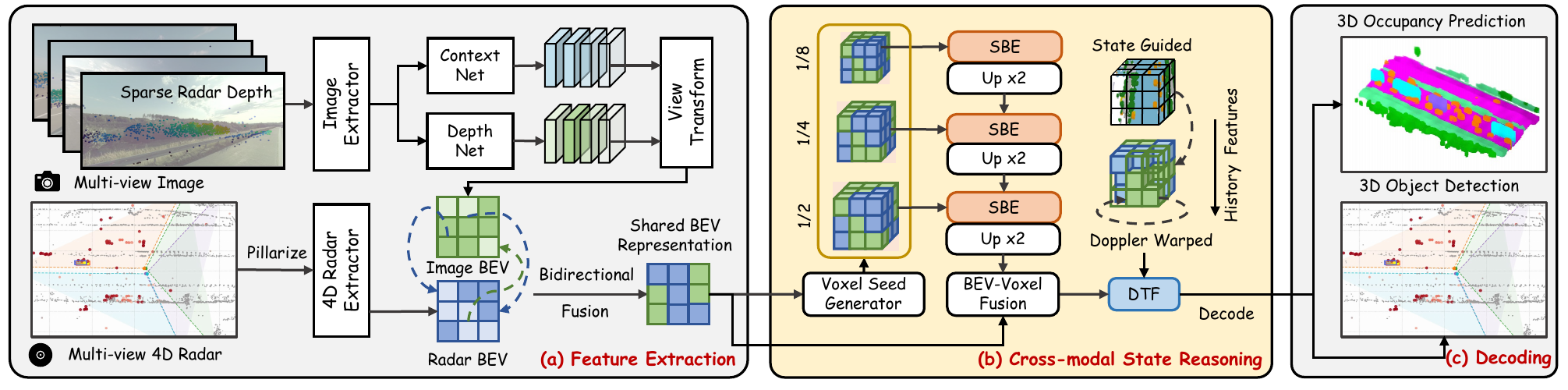}
    \vspace{-15pt}
    \caption{Overall framework of \method{}. The pipeline first
    extracts multi-view image BEV features and 4D radar BEV features, fuses them into a
    shared BEV representation, reasons over coarse-to-fine occupancy states with
    SBE and DTF, and decodes joint 3D detection and semantic occupancy outputs.}
    \label{fig_framework}
    \vspace{-5pt}
\end{figure*}
Driven by recent datasets and radar-camera perception methods, 4D
millimeter-wave radar is becoming an important sensing modality for autonomous
driving \cite{VoD,TJ4D,Kradar,RCFusion,RCBEVDet,SGDet3D,RaGS2026,R4Det2026}.
It provides stable geometric and motion cues under challenging conditions.
However, its sparse returns still require camera semantics for complete scene
understanding. Recent radar-camera methods therefore improve bird's-eye-view
fusion, semantic-geometric interaction, sparse object representation, and
temporal modeling
\cite{RCFusion,RCBEVDet,SGDet3D,RaGS2026,R4Det2026}. Nevertheless, most of these
advances are either validated on front-view detection benchmarks such as
View-of-Delft and TJ4DRadSet \cite{VoD,TJ4D}, or remain primarily optimized for
box-level perception in surround-view settings. This field gap limits how 4D
radar evidence is used for full-scene perception, where object reasoning and
scene layout should be evaluated under a unified radar-camera protocol.

Figure~\ref{fig_motivation} further shows how representative 4D radar-camera
systems have progressively expanded this scope, while a key limitation remains.
SGDet3D \cite{SGDet3D} focuses on front-view detection. RCBEVDet
\cite{RCBEVDet} extends radar-camera detection to 360{$^\circ$} perception.
Doracamom \cite{Doracamom2026} further moves to joint detection and occupancy
prediction. However, the representation form remains close to a shared BEV
feature followed by a detection head and an occupancy branch.

Yet introducing occupancy only after the shared BEV representation is formed
makes the bottleneck more than task coverage. What remains missing is a
stateful representation that lets occupancy shape object reasoning before final
decoding. Multi-task perception offers a relevant
perspective: SOGDet uses occupancy context for detection, and UniVision studies
a unified formulation for joint 3D perception
\cite{SOGDet2024,UniVision2024}. However, neither method is designed for 4D
radar-camera multimodality or radar Doppler motion cues. Existing radar-camera
systems such as RCBEVDet and BEVFusion focus on BEV fusion for detection, while
Doracamom attaches occupancy as a joint output branch after the shared feature
\cite{RCBEVDet,BEVFusion,Doracamom2026}. None of these designs expose
occupancy as an explicit intermediate reasoning state that guides feature
refinement and temporal memory before final decoding.
Evaluation is also concentrated on OmniHD-Scenes, so cross-dataset generality
remains insufficiently examined. These limitations motivate a
representation-level design that introduces occupancy as an intermediate scene
state for 4D radar-camera multi-task perception. This state links
radar-supported foreground geometry and motion cues with continuous scene
layout, allowing dense occupancy evidence and temporal cues to enter the shared
representation before final prediction.

To realize this idea, we propose \method{}, a 4D radar-camera framework for
360$^\circ$ full-scene perception. \method{} organizes the shared BEV
representation through cross-modal state reasoning. First, it converts camera
appearance and radar geometry into occupancy-aware state features before final
decoding. It then applies State-guided BEV Enhancement (SBE) to refine the
current BEV representation through state-guided deformable cross attention.
Finally, Doppler-guided Temporal Fusion (DTF) uses ego alignment and
motion-sensitive radar evidence to preserve reliable state history across
frames. Through this design, occupancy state strengthens object-level features
without replacing the detector representation. Beyond the model, we also extend
the benchmark setting beyond OmniHD-Scenes, the main multiview 4D radar-camera
dataset with joint detection and occupancy labels. Specifically, we construct
occupancy annotations for ManTruckScenes from satellite-map priors and build a
unified evaluation protocol across the two datasets. Together, these model and benchmark components lead to four contributions.
\begin{itemize}
    \item We identify semantic occupancy as a persistent scene state for
    4D radar-camera perception that unifies layout, temporal evidence, and
    object-level reasoning.
    \item We propose \method{}, a state-centered 4D radar-camera framework that
    performs cross-modal occupancy-state reasoning through SBE and DTF for feature refinement
    and temporal state modeling.
    \item We extend ManTruckScenes with satellite-map-based occupancy 
    labels and establish a unified
    multi-task protocol for advanced 4D radar-camera perception.
    \item We define comprehensive experiments on accuracy, robustness,
    and efficiency, establishing a unified protocol for dual-task
    4D radar-camera perception.
\end{itemize}

\section{Related Work}

\subsection{4D Radar-Camera 3D Object Detection}
4D radar-camera detection uses radar geometry and motion cues to strengthen
camera perception under sparse or ambiguous observations. Benchmarks such as
View-of-Delft, TJ4DRadSet, and K-Radar support this line of work
\cite{VoD,TJ4D,Kradar}. Early methods, including CenterFusion, CRAFT, and
RADIANT, attach radar evidence to image proposals or object centers
\cite{centerfusion,CRAFT,RADIANT}. Later BEV methods, including RCFusion,
RCBEVDet, LXL, SGDet3D, and HGSFusion, lift both modalities into shared BEV
features for scene-level fusion \cite{RCFusion,RCBEVDet,LXL,SGDet3D,HGSFusion}.
RaGS and R4Det further strengthen sparse representation, depth reasoning, and
temporal modeling \cite{RaGS2026,R4Det2026}. Recent extensions also explore
local-global radar detection, sparse-to-dense radar learning, raw radar-tensor
fusion, radar-camera depth estimation, cross-view instance awareness, and
collaborative perception
\cite{LGDD2025,SD4R2026,WRCFormer2025,SARCD2025,SIFormer2026,RCGeoCP2026}.

These methods establish strong radar-camera detection baselines and mark the
progression from front-view detection to 360{$^\circ$} perception. However,
their objective remains box-centric: radar evidence is fused for detection,
whereas dense scene structure is rarely maintained as a reusable state for joint
object and occupancy reasoning.

\subsection{Multi-Task Learning for Perception}
Multi-task perception seeks shared representations that support multiple
driving outputs, with semantic occupancy providing a dense description of
scene layout. Camera BEV and
occupancy baselines such as MonoScene, VoxFormer, SurroundOcc,
PanoOcc, and M-CONet from OpenOccupancy recover voxelized layout from
multi-view imagery
\cite{MonoScene,VoxFormer,SurroundOcc,wang2024panoocc,wang2023openoccupancy,Gaussianformer}.
SOGDet and UniVision further show that
occupancy context can support object reasoning
\cite{SOGDet2024,UniVision2024}. These methods motivate joint perception, but
they do not target 4D radar-camera perception, where sparse, motion-aware
radar returns are naturally sensitive to foreground objects.

Benchmark support for 4D radar-camera multi-task perception is much narrower.
OmniHD-Scenes is currently the only public 4D radar-camera benchmark with both
3D detection and semantic occupancy annotations \cite{OmniHD}, and Doracamom
provides a joint reference on this setting \cite{Doracamom2026}. This leaves
cross-dataset generality underexplored and ties future radar-aware end-to-end
studies to one benchmark contract. We therefore extend ManTruckScenes with
auditable occupancy labels and evaluate both datasets under one normalized
radar-camera detection-and-occupancy protocol. Methodologically, we further
differ from ordinary dual-head systems by propagating occupancy as an explicit
scene state, together with Doppler motion cues, inside the shared
representation rather than decoding it only at the output side.

\section{Method}
\subsection{Overview}
\method{} formulates 4D radar-camera perception as cross-modal state reasoning,
as summarized in Fig.~\ref{fig_framework}. Given synchronized multi-view images
and 4D radar sweeps at time \(t\), Fig.~\ref{fig_framework} (a) constructs an
image BEV through view transformation and fuses it with radar BEV using
bidirectional deformable attention. The fused radar-camera BEV is then lifted
into a coarse-to-fine voxel hierarchy, which carries the scene states refined in
Fig.~\ref{fig_framework} (b). SBE injects current-frame occupancy structure,
while DTF propagates temporally aligned state evidence with radar-motion cues.
The detection and occupancy heads then read out from the shared state-refined
representation, as shown in Fig.~\ref{fig_framework} (c). Implementation details
are provided in the supplementary materials.

\begin{figure}[t]
    \centering
    \includegraphics[width=1.0\linewidth]{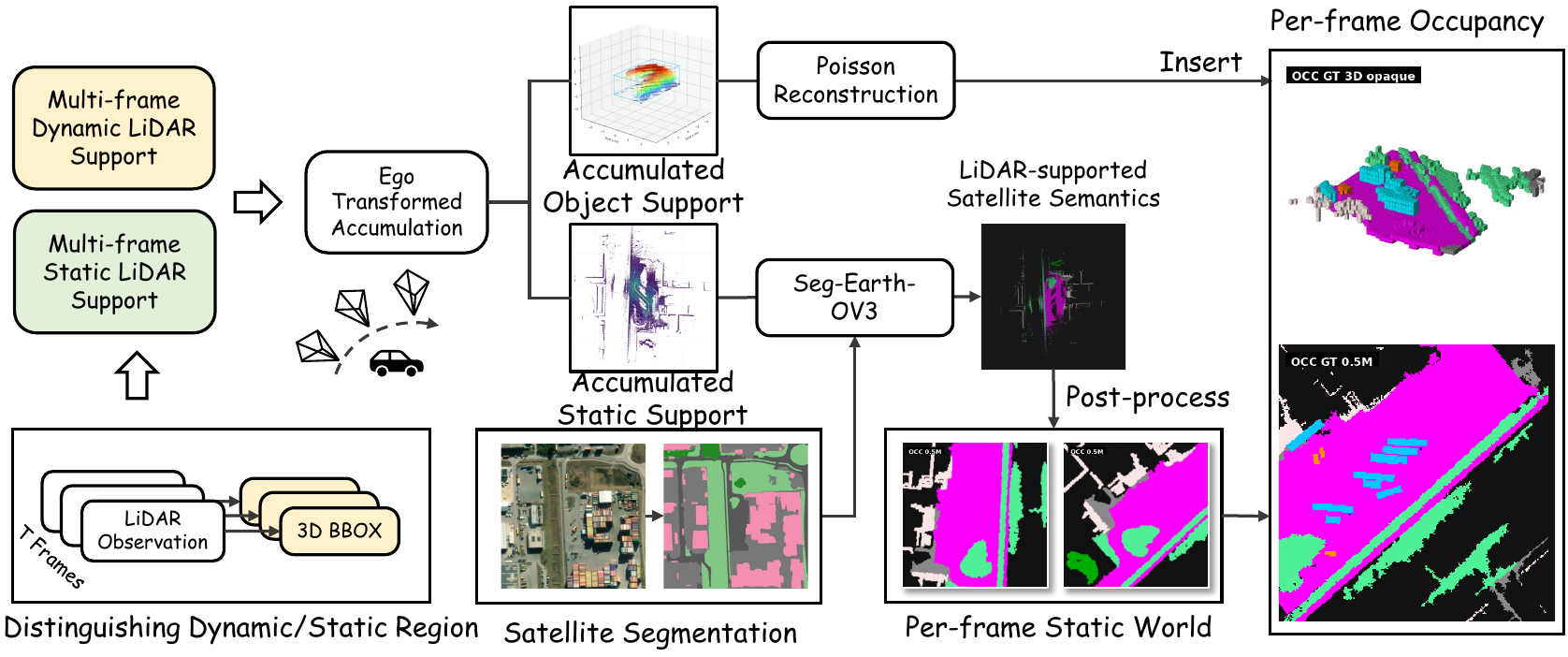}
    \caption{ManTruckScenes occupancy construction.
    LiDAR anchors static geometry, satellite priors
    provide conservative static semantics, and
    object-local replay inserts dynamic surfaces,
    which together enable auditable cross-dataset
    occupancy evaluation.}
    \vspace{-10pt}
    \label{fig:mantruck_occ}
\end{figure}

\subsection{ManTruckScenes Occupancy Construction}
\label{subsec:benchmark_construction}
A single dataset provides limited evidence for full-scene radar-camera
evaluation. OmniHD-Scenes provides native multiview 4D radar-camera detection
and occupancy labels, but it is currently the only benchmark of this type.
ManTruckScenes offers a complementary commercial-vehicle platform, yet it
contains detection annotations only. Evaluating it only as a detector benchmark
would leave the occupancy part of radar-camera full-scene perception untested
across datasets. We therefore extend ManTruckScenes with an auditable occupancy
export.

As illustrated in Fig.~\ref{fig:mantruck_occ}, we first aggregate multi-frame
LiDAR observations into a static support world, so occupied geometry is
anchored by measured surfaces. Satellite segmentation is then used only for
conservative static semantics around supported regions, while dynamic actors
are inserted by object-local surface replay instead of dense box filling. The
resulting labels are used only as training and evaluation targets, with fixed
label mapping, voxel range, masks, temporal metadata, and metrics within each
dataset. The supplementary material provides the full generation and
quality-control details.

\subsection{Feature Extractor}
The feature extractor in Fig.~\ref{fig_framework} (a) builds the radar-camera
BEV evidence used to initialize the state. We omit the batch dimension. Given
synchronized images from \(N\) cameras, a ResNet-FPN encoder extracts
\(\mathbf{F}^{2D}_t\in\mathbb{R}^{N\times C_i\times H_i\times W_i}\). A
depth-aware view transformer produces the camera BEV feature, depth
distribution, and image-view context:
\begin{equation}
    (\mathbf{B}^{c}_t,\mathbf{D}_t,\mathbf{F}^{ctx}_t) =
    \mathcal{P}_{view}(\mathbf{F}^{2D}_t,\Pi_t),
\end{equation}
where \(\Pi_t\) collects camera calibration and image augmentations. The
outputs are image BEV
\(\mathbf{B}^{c}_t\in\mathbb{R}^{C\times H\times W}\), depth
\(\mathbf{D}_t\in\mathbb{R}^{N\times D\times H_i\times W_i}\), and multi-view
context
\(\mathbf{F}^{ctx}_t\in\mathbb{R}^{N\times C\times H_i\times W_i}\).
Following \cite{SGDet3D}, we also use a geometric projection of
\(\mathbf{F}^{ctx}_t\) to strengthen the camera BEV.

The radar branch maps the synchronized 4D radar point set
\(\mathcal{R}_t=\{(x,y,z,\mathbf{a})\}\) to
\(\mathbf{B}^{r}_t\in\mathbb{R}^{C_r\times H\times W}\) with a pillarization
encoder, where \(\mathbf{a}\) contains Doppler and other radar attributes. The
pillarization details are provided in the supplementary materials. Image and
radar BEV features are then fused by bidirectional deformable BEV attention
following \cite{RCBEVDet}, producing \(\mathbf{B}^{rc}_t\). This shared BEV
representation aligns camera layout cues with radar foreground evidence.

Finally, each pyramid level of \(\mathbf{B}^{rc}_t\) is channel-adapted and
lifted along height by a learned height MLP:
\begin{equation}
    \mathbf{V}^{rc,s}_t =
    \operatorname{Reshape}\big(
    \mathcal{H}_{s}(\operatorname{Down}_{s}(\mathbf{B}^{rc}_t))\big),
\end{equation}
where
\(\mathbf{V}^{rc,s}_t\in
\mathbb{R}^{C_s\times Z_s\times H_s\times W_s}\). This produces multi-scale
fused voxel features \(\{\mathbf{V}^{rc,s}_t\}_{s=0}^{S}\), which provide the
substrate for the following state reasoning modules.

\begin{figure}[t]
    \centering
    \includegraphics[pagebox=cropbox,width=1.0\linewidth]{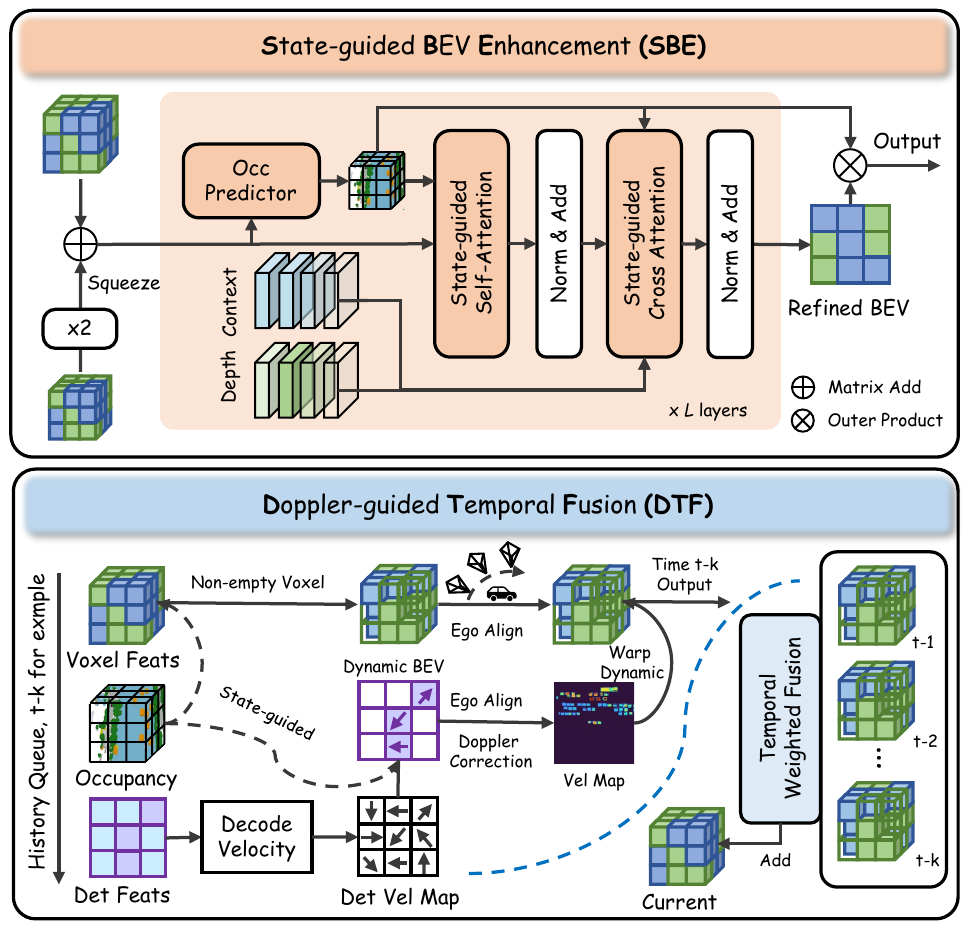}
    \caption{State reasoning modules. SBE predicts occupancy
    state before attention and uses non-empty confidence with depth-aware image
    context to refine features. DTF decodes velocity from detection features,
    uses occupancy confidence as non-empty support, applies Doppler correction,
    and performs dynamic warping and temporal weighted fusion.}
    \vspace{-10pt}
    \label{fig_sbe_dtf}
\end{figure}

\subsection{Cross-modal State Representation}
At stage \(s\), \method{} represents the scene with a voxel
feature \(\mathbf{V}^{rc,s}_t\) and an occupancy-logit state
\(\mathbf{O}^{s}_t\in\mathbb{R}^{H_s\times W_s\times Z_s\times C_{occ}}\). The
empty channel defines a non-empty confidence map
\(\mathcal{M}(\mathbf{O})=1-\operatorname{softmax}(\mathbf{O})_{[...,e]}\),
where \(e\) denotes the empty class and \(\mathbf{O}\) always denotes logits.
\(\mathcal{M}(\mathbf{O})\) is broadcast over channels on voxel features and
height-pooled when applied to BEV features. This operator is the interface
between occupancy prediction and feature reasoning, so the state is not a
terminal side output. It is predicted before attention, resized across stages as
a semantic prior, and converted back into voxel features for both final heads.
We summarize the stage transition as
\begin{equation}
    (\mathbf{O}^{s}_t,\widetilde{\mathbf{V}}^{rc,s}_t)
    =
    \Phi^{s}_{state}(\mathbf{V}^{rc,s}_t,\mathbf{F}^{ctx}_t,
    \mathbf{D}_t,\Pi_t,\mathbf{O}^{s-1}_t),
\end{equation}
where \(\Phi^{s}_{state}\) includes state prediction, SBE, and
occupancy-aware voxel recovery. SBE internally produces the stage BEV feature
\(\mathbf{B}^{rc,s}_t\), which is lifted back into the recovered voxel state.
The recovered \(\widetilde{\mathbf{V}}^{rc,s}_t\) is upsampled as residual state
evidence for the next finer scale.

\subsection{State-guided BEV Enhancement (SBE)}
\label{subsec:SBE}

Given the state interface above, SBE converts the predicted occupancy state into
current-frame spatial guidance for radar-camera scene understanding. It is the
stage-wise spatial decoder in the upper panel of Fig.~\ref{fig_sbe_dtf}. At
stage \(s\), it takes the voxel seed
\(\mathbf{V}^{rc,s}_t\), image-view context \(\mathbf{F}^{ctx}_t\), depth
distribution \(\mathbf{D}_t\), camera geometry \(\Pi_t\), and the previous
coarser state \(\mathbf{O}^{s-1}_t\) when available. An occupancy predictor first
estimates \(\mathbf{O}^{s}_t\), from which
\(\mathcal{M}(\mathbf{O}^{s}_t)\) is obtained.

The predicted state guides two attention operations. State-guided self-attention
uses non-empty confidence to bias BEV query offsets and weights, so foreground
and uncertain occupied regions receive more modeling capacity. State-guided
cross-attention then projects height anchors to camera views and uses depth
consistency and non-empty confidence to weight the sampled context. For a BEV
location \((x,y)\), let \(\mathcal{X}_{x,y}\) be its height anchors. The image
value sampled at anchor \(\mathbf{x}\) and view \(j\) is
\begin{equation}
    \mathbf{g}_{j\mathbf{x}} =
    \mathbf{W}_{v}
    \mathtt{Bilinear}\big(\mathbf{F}^{ctx}_{t,j},
    \mathcal{P}(\mathbf{x},\Pi_{t,j})
    +\Delta\mathbf{u}_{j\mathbf{x}}\big),
\end{equation}
and the state-guided aggregation is
\begin{equation}
    \bar{\mathbf{b}}^{rc,s}_t(x,y) =
    \sum_{j=1}^{N}
    \sum_{\mathbf{x}\in\mathcal{X}_{x,y}}
    \alpha_{j\mathbf{x}}d_{j\mathbf{x}}
    \mathcal{M}(\mathbf{O}^{s}_t)(\mathbf{x})\,
    \mathbf{g}_{j\mathbf{x}},
\end{equation}
where \(\alpha_{j\mathbf{x}}\) is the deformable attention weight,
\(d_{j\mathbf{x}}\) is obtained by interpolating \(\mathbf{D}_{t,j}\) at the
projected pixel and anchor-depth bin, \(\Delta\mathbf{u}_{j\mathbf{x}}\) is the
learned image-plane offset, and \(\mathcal{M}(\mathbf{O}^{s}_t)(\mathbf{x})\)
provides a continuous non-empty confidence rather than a hard occupied-anchor
mask.

\begin{table*}[!htbp]
    \centering
    \footnotesize
    \belowrulesep=0pt
    \aboverulesep=0pt
    \setlength{\tabcolsep}{1pt}
    \renewcommand\arraystretch{1.00}
    \begin{tabular}{@{}C{5.0cm}|C{0.8cm}|C{1.0cm}C{1.0cm}|C{1.0cm}C{1.0cm}C{1.0cm}C{1.0cm}|C{0.96cm}C{0.96cm}C{0.96cm}C{0.96cm}|C{0.96cm}@{}}
    \toprule[1.0pt]
    \detmethod{Methods} & \modbox{Mod.} & \detbox{mAP$\uparrow$} &
    \detbox{ODS$\uparrow$} & \detbox{mATE$\downarrow$} &
    \detbox{mASE$\downarrow$} & \detbox{mAOE$\downarrow$} &
    \detbox{mAVE$\downarrow$} & \detclsbox{Car$\uparrow$} &
    \detclsbox{Ped.$\uparrow$} & \detclsbox{Rider$\uparrow$} &
    \detclsbox{LVeh.$\uparrow$} & \detclsbox{FPS$\uparrow$}\\
    \tabgap
    \hline
    \tabgap
    PointPillars (CVPR 2019) & R & 23.82 & 37.21 &
    0.6752 & 0.2447 & 0.3776 & 0.6789 & 52.74 & 0.69 & 28.57 & 13.29 & \bestcell{62.2}\\
    RadarPillarNet (IEEE TIM 2023) & R & 24.88 & 37.81 &
    0.6597 & 0.2389 & 0.3736 & 0.6982 & 52.99 & 2.06 & 29.45 & 15.02 & \secondcell{60.3}\\
    \tabgap
    \hline
    BEVFormer (ECCV 2022) & C & 29.17 &
    30.54 & 1.1046 & 0.2346 & 0.4889 & 1.0797 & 53.64 & 14.48 & 33.55 & 15.01 &
    \thirdcell{11.4}\\
    PanoOcc (CVPR 2024) & C & 29.17
    & 28.55 & 1.1500 & 0.2446 & 0.6378 & 1.6066 & 51.58 & 15.82 & 35.02 & 14.26 &
    5.5\\
    \tabgap
    \hline
    \tabgap
    BEVFusion (NeurIPS 2022) & C\&R &
    33.95 & 42.62 & 0.5730 & 0.2465 & 0.3814 & 0.7474 & 56.25 & 11.66
    & 50.90 & 16.99 & 3.6\\
    RCFusion (IEEE TIM 2023) & C\&R &
    34.88 & 40.65 & 0.5676 & 0.2535 & 0.4011 & 0.9208 & 57.17
    & 12.87 & 51.35 & 18.11 & 3.6\\
    RCBEVDet (CVPR 2024) & C\&R & 35.53 &
    45.04 & \underline{0.5138} & \textbf{0.2305} & 0.3914 &
    0.6825 & \thirdcell{62.35} & 10.11 & \thirdcell{54.60} & 15.06 & 5.2\\
    SGDet3D (RAL 2025) & C\&R & \secondcell{41.73} &
    \secondcell{47.70} & 0.5430 & 0.2369 & 0.3885 & 0.6849 &
    \secondcell{63.30} & \thirdcell{19.00} & \secondcell{58.30} & \secondcell{26.30} &
    3.4 \\
    RaGS (CVPR 2026) & C\&R &
    35.88 & 43.45 & -- & -- & -- & -- &
    -- & -- & -- & -- & --\\
    Doracamom-S (TCSVT 2026) & C\&R & 37.60 &
    41.31 & 0.6724 & \underline{0.2329} & 0.4359 & 0.8579 &
    58.94 & 17.84 & 52.72 & 20.89 &
    4.8\\
    Doracamom (TCSVT 2026) & C\&R & \thirdcell{39.12} &
    \thirdcell{46.22} & 0.6646 & 0.2331 & \textbf{0.3545} & \underline{0.6151} &
    61.12 & \secondcell{19.83} & 53.35 & \thirdcell{22.18} &
    4.2\\
    \cellcolor{gray!20}\method{}~\textbf{(ours)} & \cellcolor{gray!20}C\&R &
    \bestcell{45.05} & \bestcell{51.40} & \textbf{0.4705} &
    0.2423 & \underline{0.3762} & \textbf{0.6013} &
    \bestcell{65.47} & \bestcell{21.73} & \bestcell{60.41} &
    \bestcell{32.58} & 4.5\\
    \tabgap
    \bottomrule[1.0pt]
    \end{tabular}
    \vspace{-5pt}
    \caption{OmniHD-Scenes 3D detection results.}
    \label{tab_omnihd_det}
    \vspace{-5pt}
\end{table*}

\begin{table*}[t]
\centering
\vspace{-5pt}
\footnotesize 
\belowrulesep=0pt
\aboverulesep=0pt
\setlength{\tabcolsep}{1pt}
\renewcommand\arraystretch{1.00}
\begin{tabular}{@{}C{5.0cm}|C{0.8cm}|C{1.0cm}C{1.0cm}|C{1.0cm}C{1.0cm}C{1.0cm}C{1.0cm}|C{0.96cm}C{0.96cm}C{0.96cm}C{0.96cm}|C{0.96cm}@{}}
\toprule[1.0pt]
\detmethod{Methods} & \modbox{Mod.} & \detbox{mAP$\uparrow$} &
\detbox{NDS$\uparrow$} & \detbox{mATE$\downarrow$} &
\detbox{mASE$\downarrow$} & \detbox{mAOE$\downarrow$} &
\detbox{mAVE$\downarrow$} & \detclsbox{Car$\uparrow$} &
\detclsbox{LVeh.$\uparrow$} & \detclsbox{Trailer$\uparrow$} &
\detclsbox{Obs.$\uparrow$} & \detclsbox{FPS$\uparrow$}\\
\tabgap
\hline
\tabgap
RadarPillarNet (IEEE TIM 2023) & R &
26.61 & 37.73 & 0.5243 & 0.2413 &
0.1689 & 2.1915 & 44.43 & 30.66 &
28.65 & 2.68 & \bestcell{64.2}\\
LGDD (IROS 2025) & R &
29.83 & 40.49 & 0.4922 & 0.2328 &
0.1229 & 1.5775 & 48.90 & 36.20 &
32.40 & 1.80 & \secondcell{20.1}\\
\tabgap
\hline
BEVFormer (ECCV 2022)& C &
28.30 & 35.47 & 0.7662 & 0.2583 &
0.1982 & 1.2747 & 39.56 & 28.10 &
19.66 & 25.87 & \thirdcell{10.7}\\
PanoOcc (CVPR 2024)& C &
28.82 & 35.72 & 0.7686 & 0.2653 &
0.1920 & 1.2623 & 40.12 & 28.22 &
21.73 & 25.20 & 6.3\\
\hline
\tabgap
BEVFusion (NeurIPS 2022) & C\&R &
30.97 & 40.65 & 0.5168 & 0.2486 &
0.1249 & 2.1377 & 47.24 & 36.09 &
30.15 & 10.40 & 4.6\\
RCFusion (IEEE TIM 2023) & C\&R &
31.61 & 40.57 & 0.5652 & 0.2483 &
\textbf{0.1155} & 2.0670 & 48.51 & 36.84 &
28.58 & 12.49 & 5.4\\
LXL (IEEE TIV 2024) & C\&R &
34.70 & 43.04 & 0.4972 & 0.2470 &
\underline{0.1174} & 1.9877 & 51.95 & 41.14 &
31.41 & 14.30 & 5.2\\
RCBEVDet (CVPR 2024) & C\&R &
41.72 & 46.85 & 0.4500 & 0.2658 &
0.1543 & \underline{1.0407} & \bestcell{61.93} & 38.00 &
34.38 & \thirdcell{32.58} & 6.5\\
HGSFusion (AAAI 2025)& C\&R &
\secondcell{43.90} & \secondcell{48.23} & 0.4607 & \underline{0.2272} &
0.1668 & 1.9479 & 58.76 & \thirdcell{43.64} &
\secondcell{35.43} & \secondcell{37.77} & 5.1\\
SGDet3D (RAL 2025)& C\&R &
41.37 & \thirdcell{47.50} & \underline{0.4252} & \textbf{0.2159} &
0.1523 & 1.6112 & \thirdcell{60.82} & \bestcell{47.38} &
26.94 & 30.29 & 3.0\\
Doracamom (TCSVT 2026) & C\&R &
\thirdcell{41.98} & 46.36 & 0.4874 & 0.2411 &
0.1982 & 1.5566 & \secondcell{61.62} & 42.71 &
\thirdcell{34.45} & 29.15 & 5.2\\
\cellcolor{gray!20}\method{}~\textbf{(ours)} & \cellcolor{gray!20}C\&R &
\bestcell{49.57} & \bestcell{52.97} & \textbf{0.3613} & 0.2485 &
0.1571 & \textbf{0.9442} & 59.41 & \secondcell{44.71} &
\bestcell{41.20} & \bestcell{52.97} & 5.5\\
\tabgap
\bottomrule[1.0pt]
\end{tabular}
\vspace{-5pt}
\caption{ManTruckScenes 3D detection results.}
\label{tab_mantruck_det}
\vspace{-15pt}
\end{table*}

The aggregated response is added to the stage BEV query and refined by a
lightweight convolutional block, yielding \(\mathbf{B}^{rc,s}_t\). The BEV
feature is then lifted back to the voxel lattice. Let
\(\mathcal{L}_{s}(\mathbf{B}^{rc,s}_t)\) denote the height-lifted stage BEV
feature. The recovered voxel state is
\begin{equation}
\begin{aligned}
    \widetilde{\mathbf{V}}^{rc,s}_t
    =
    \mathbf{V}^{rc,s}_t
    &+\mathcal{L}_{s}(\mathbf{B}^{rc,s}_t)
    \odot \mathcal{M}(\mathbf{O}^{s}_t),
\end{aligned}
\end{equation}
where the second term gives explicit non-empty weighting. The recovered voxel
feature is upsampled and passed to the next finer stage. Multi-scale occupancy
supervision is applied to these stage states, while detailed SBE variants are
reported in the supplementary materials.

\begin{table*}[htbp]
    \centering
    \vspace{-5pt}
    \begin{threeparttable}
    \footnotesize
    \belowrulesep=0pt
    \aboverulesep=0pt
    \setlength{\tabcolsep}{1pt}
    \renewcommand\arraystretch{1.00}
    \begin{tabular}{c|c|cc|ccccccccccc}
    \toprule[1.0pt]
    \occmethod{Methods} & \modbox{Mod.} & \occbox{SC IoU} &
    \occbox{mIoU} &
    \occbox{\rotatebox{90}{\textcolor{ncar}{$\blacksquare$}Car}} &
    \occbox{\rotatebox{90}{\textcolor{npedestrian}{$\blacksquare$}Ped.}} &
    \occbox{\rotatebox{90}{\textcolor{nlarge_vehicle}{$\blacksquare$}LVeh.}} &
    \occbox{\rotatebox{90}{\textcolor{ncycle}{$\blacksquare$}Cyc.}} &
    \occbox{\rotatebox{90}{\textcolor{nroad_obstacle}{$\blacksquare$}Obs.}} &
    \occbox{\rotatebox{90}{\textcolor{ntraffic_fence}{$\blacksquare$}Fence}} &
    \occbox{\rotatebox{90}{\textcolor{ndriveable_surface}{$\blacksquare$}Drive}} &
    \occbox{\rotatebox{90}{\textcolor{nvegetation}{$\blacksquare$}Veg.}} &
    \occbox{\rotatebox{90}{\textcolor{nmanmade}{$\blacksquare$}Man.}} &
    \occbox{\rotatebox{90}{\textcolor{nrider}{$\blacksquare$}Rider}} &
    \occbox{\rotatebox{90}{\textcolor{nsidewalk}{$\blacksquare$}Side.}} \\
    \tabgap
    \hline
    \tabgap
    BEVFormer (ECCV 2022) & C & 28.42 &
    16.23 & 22.73 & 5.45 & 18.21 & 3.09 & 3.87 & 21.54 & 48.15 & 17.77 &
    5.48 & 14.70 & \thirdcell{17.58} \\
    PanoOcc (CVPR 2024) & C
    & 26.36 & 15.20 & 22.42 & 5.91 & 17.98 & 3.11 & 3.36 & 21.46 &
    \thirdcell{50.47} & 11.20 & 1.80 & 13.58 & 15.90 \\
    \tabgap
    \hline
    \tabgap
    BEVFusion (NeurIPS 2022) & C\&R &
    27.02 & 16.24 & 27.02 & 4.78 & 21.59 & 1.55 & 2.78 & 25.21 & 44.35 &
    13.06 & 4.25 & 21.71 & 12.32 \\
    M-CONet (ICCV 2023) & C\&R &
    27.74 & 16.08 & 25.21 & 3.42 & 21.46 & 0.88 & 0.58 &
    29.88 & 34.48 & 19.57 & 8.98 & 17.53 &
    14.89 \\
    SGDet3D (RAL 2025) & C\&R &
    31.15 & \thirdcell{21.19} & 28.74 & \secondcell{10.35} &
    21.77 & \secondcell{5.92} & \secondcell{9.02} &
    \secondcell{36.13} & 47.90 & \thirdcell{20.72} &
    \secondcell{12.12} & \secondcell{24.45} & 15.96 \\
    Doracamom-S (TCSVT 2026) & C\&R &
    \thirdcell{31.46} & 19.49 & \thirdcell{30.10} & 6.71 & \thirdcell{24.31} &
    2.85 & 6.55 & 25.77 & 49.72 &
    19.57 & 8.72 & 23.60 & 16.53 \\
    Doracamom (TCSVT 2026) & C\&R &
    \secondcell{33.96} & \secondcell{21.81} & \secondcell{30.81} & \thirdcell{7.22} & \secondcell{24.70} &
    \thirdcell{4.49} & \thirdcell{7.84} & \thirdcell{34.49} & \secondcell{52.00} &
    \secondcell{21.68} & \thirdcell{11.49} & \thirdcell{24.33} & \secondcell{20.86} \\
    \cellcolor{gray!20}\method{}~\textbf{(ours)} & \cellcolor{gray!20}C\&R &
    \bestcell{35.05} & \bestcell{24.71} &
    \bestcell{33.71} & \bestcell{11.29} &
    \bestcell{27.41} & \bestcell{6.50} &
    \bestcell{10.19} & \bestcell{41.98} &
    \bestcell{52.52} & \bestcell{22.38} &
    \bestcell{14.52} & \bestcell{29.75} &
    \bestcell{21.54} \\
    \tabgap
    \hline
    \end{tabular}
    \end{threeparttable}
    \vspace{-5pt}
    \caption{OmniHD-Scenes semantic occupancy results.}
    \label{tab_omnihd_occ}
    \vspace{-5pt}
\end{table*}

\begin{table*}[t]
\centering
\vspace{-5pt}
\footnotesize
\belowrulesep=0pt
\aboverulesep=0pt
\setlength{\tabcolsep}{1pt}
\renewcommand\arraystretch{1.00}
\begin{tabular}{c|c|cc|cccccccccc}
\toprule[1.0pt]
\occmethod{Methods} & \modbox{Mod.} & \occbox{SC IoU} &
\occbox{mIoU} &
\mtoccbox{\rotatebox{90}{\textcolor{ncar}{$\blacksquare$}Car}} &
\mtoccbox{\rotatebox{90}{\textcolor{nlarge_vehicle}{$\blacksquare$}LVeh.}} &
\mtoccbox{\rotatebox{90}{\textcolor{ntrailer}{$\blacksquare$}Trailer}} &
\mtoccbox{\rotatebox{90}{\textcolor{nego_trailer}{$\blacksquare$}EgoTr.}} &
\mtoccbox{\rotatebox{90}{\textcolor{nroad_obstacle}{$\blacksquare$}Obs.}} &
\mtoccbox{\rotatebox{90}{\textcolor{ndriveable_surface}{$\blacksquare$}Drive}} &
\mtoccbox{\rotatebox{90}{\textcolor{notherflat}{$\blacksquare$}O.Flat}} &
\mtoccbox{\rotatebox{90}{\textcolor{nterrain}{$\blacksquare$}Terr.}} &
\mtoccbox{\rotatebox{90}{\textcolor{nvegetation}{$\blacksquare$}Veg.}} &
\mtoccbox{\rotatebox{90}{\textcolor{nmanmade}{$\blacksquare$}Man.}} \\
\tabgap
\hline
BEVFormer (ECCV 2022) & C &
21.04 & 23.35 & 27.74 & 29.06 &
36.22 & 76.22 & 10.76 & 22.41 &
10.65 & 12.37 & 5.13 & 2.94\\
PanoOcc (CVPR 2024) & C &
22.75 & 24.43 & 28.68 & 29.51 &
37.31 & \thirdcell{76.21} & 11.16 & 25.26 &
12.32 & 13.31 & 6.21 & 4.33\\
\hline
\tabgap
BEVFusion (NeurIPS 2022) & C\&R &
\thirdcell{24.74} & 24.69 & 31.68 & 31.68 &
38.89 & 75.37 & 5.84 & 24.30 &
12.95 & 15.14 & 9.59 & 1.46\\
RCFusion (IEEE TIM 2023) & C\&R &
21.91 & 26.02 & 35.95 & 35.11 &
47.18 & 75.47 & 6.26 & 19.24 &
11.99 & 14.97 & 9.54 & 4.44\\
LXL (IEEE TIV 2024) & C\&R &
22.73 & 27.92 & 37.18 & 38.14 &
47.64 & 76.39 & 9.87 & \thirdcell{25.78} &
12.61 & \thirdcell{18.22} & 10.16 & 3.21\\
RCBEVDet (CVPR 2024)  & C\&R &
24.02 & 28.69 & 40.39 & 37.75 &
48.94 & 76.62 & \secondcell{18.99} & 22.61 &
13.08 & 14.74 & 9.57 & 4.21\\
HGSFusion (AAAI 2025) & C\&R &
23.87 & \secondcell{29.36} & \thirdcell{42.10} & \thirdcell{40.86} &
\secondcell{51.62} & \secondcell{76.69} & \thirdcell{17.30} & 21.32 &
12.09 & 15.81 & 10.49 & \secondcell{5.30}\\
SGDet3D (RAL 2025) & C\&R &
24.48 & \thirdcell{28.74} & 36.63 &
\secondcell{42.22} & 50.26 & 66.00 &
15.87 & \secondcell{27.34} & \thirdcell{13.14} &
\secondcell{18.42} & \thirdcell{11.36} & \bestcell{6.13}\\
Doracamom (TCSVT 2026) & C\&R &
\secondcell{24.93} & 28.65 & \secondcell{42.59} &
37.58 & \thirdcell{50.45} & 76.56 &
3.28 & \bestcell{28.09} & \secondcell{13.50} &
18.05 & \secondcell{11.55} & \thirdcell{4.85}\\
\cellcolor{gray!20}\method{}~\textbf{(ours)} & \cellcolor{gray!20}C\&R &
\bestcell{25.07} & \bestcell{31.65} & \bestcell{43.61} &
\bestcell{43.56} & \bestcell{53.90} & \bestcell{77.65} &
\bestcell{25.58} & 24.41 & \bestcell{13.85} &
\bestcell{18.75} & \bestcell{11.86} & 3.33\\
\tabgap
\hline
\end{tabular}
\caption{ManTruckScenes semantic occupancy results.}
\label{tab_mantruck_occ_export}
\vspace{-15pt}
\end{table*}
\begin{figure*}[t]
    \centering
    \includegraphics[width=\linewidth]{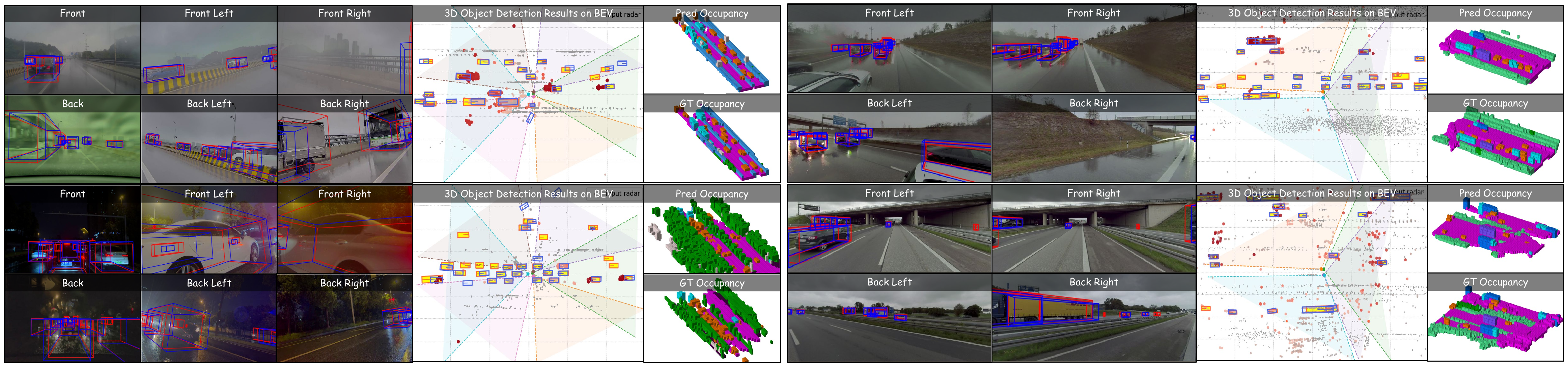}
    \vspace{-15pt}
    \caption{Qualitative results on OmniHD-Scenes (left) and ManTruckScenes (right).}
    \label{fig_visualization}
    \vspace{-15pt}
\end{figure*}
\subsection{Doppler-guided Temporal Fusion (DTF)}
\label{subsec:DTF}

DTF is the temporal interface of the final voxel state, as shown in the lower
panel of Fig.~\ref{fig_sbe_dtf}. The lower panel implements a compact
motion-conditioning path: detection features decode a BEV velocity map, the
occupancy state indexes non-empty dynamic support, and radar Doppler corrects
the motion cue before dynamic warping. Let \(\mathbf{V}_t\) be the final
occupancy-state feature and \(\mathbf{O}_t\) be its occupancy logits. We use
\(\mathcal{I}(\mathbf{O},\cdot)\) to denote state-indexed feature selection,
which suppresses empty or irrelevant support before temporal retrieval. For a
history horizon \(T_h\), DTF caches
\(\mathcal{H}_t=\{(\mathbf{V}_{t-k},\mathbf{O}_{t-k},
\mathbf{T}_{t-k},\mathbf{U}_{t-k})\}_{k=1}^{T_h}\), where
\(\mathbf{T}_{t-k}\) is the ego pose and \(\mathbf{U}_{t-k}\) is the BEV motion
field derived from box velocity and radar Doppler cues. Its input-output
contract is
\begin{equation}
\begin{aligned}
    \widetilde{\mathbf{V}}_t =
    \mathcal{F}_{DTF}\big(
    \mathbf{V}_t,\mathcal{H}_t\big)
\end{aligned}
\end{equation}
where DTF aligns historical state features and motion fields to the current ego
frame.

DTF performs state-aware history retrieval in three steps. First,
the state index \(\mathcal{I}(\mathbf{O}_{t-k}, \cdot)\) provides the non-empty
voxel path in Fig.~\ref{fig_sbe_dtf}, and the selected historical features are
ego-aligned:
\begin{equation}
    \widehat{\mathbf{V}}_{t,k}=
    \mathcal{T}_{t-k\rightarrow t}(\mathcal{I}(\mathbf{O}_{t-k}, \mathbf{V}_{t-k})).
\end{equation}
Second, the Doppler-corrected velocity map drives dynamic warping after ego
alignment. The motion field \(\mathbf{U}_{t-k}\) follows the lower path in
Fig.~\ref{fig_sbe_dtf}: detection features decode velocity, occupancy confidence
selects dynamic foreground support, and radar Doppler corrects the motion cue.
With interval \(\Delta t_k\), the displacement field is
\begin{equation}
\begin{aligned}
    \Delta \mathbf{p}_{t,k}
    =
    \operatorname{clip}\big(
    \Delta t_k
    \mathcal{T}_{t-k\rightarrow t}
    (\mathcal{I}(\mathbf{O}_{t-k}, \mathbf{U}_{t-k})),\tau\big),
\end{aligned}
\end{equation}
where \(\tau\) limits unrealistically large shifts. The ego-aligned state is
sampled from the source location that moves into each current BEV cell:
\begin{equation}
    \bar{\mathbf{V}}_{t,k} =
    \mathcal{S}(\widehat{\mathbf{V}}_{t,k},-\Delta\mathbf{p}_{t,k}),
\end{equation}
where \(\mathcal{S}\) denotes BEV-plane bilinear sampling shared by all height
bins. Indexing removes unsupported regions, but the reliability of the remaining
memory still varies across voxels. We therefore derive
\(\bar{\mathbf{C}}_{t,k}\) by applying the same ego alignment and dynamic warp to
\(\mathcal{M}(\mathbf{O}_{t-k})\). Third, temporal fusion averages the warped
memories with occupancy confidence and temporal decay:
\begin{equation}
    \mathbf{M}_t =
    \frac{\sum_{k=1}^{T_h}\lambda^k
    \bar{\mathbf{C}}_{t,k}\odot\bar{\mathbf{V}}_{t,k}}
    {\sum_{k=1}^{T_h}\lambda^k\bar{\mathbf{C}}_{t,k}+\epsilon}.
\end{equation}
The current state is then refined by a lightweight 3D convolutional update,
$
    \widetilde{\mathbf{V}}_t
    =
    \mathbf{V}_t+
    \mathtt{Conv}_{3D}\big(\mathbf{M}_t\big).
$
The same Doppler-aligned update is height-pooled and adapted to the detection
BEV, so detection and occupancy receive a shared radar-guided temporal state.
Compared with generic BEV temporal aggregation, DTF separates three measurable
factors: temporal horizon, state-aware memory retrieval, and radar-conditioned
motion compensation.

\subsection{Training Objective}
Training supervises both terminal predictions and intermediate state. We
therefore use a shared occupancy loss for the state branch and the final occupancy
branch. At stage \(s\), the occupancy loss is
$
    \mathcal{L}^{s}_{occ} =
    \mathcal{L}^{s}_{focal} +
    \mathcal{L}^{s}_{geo} + \mathcal{L}^{s}_{sem},
$
where the three terms denote weighted focal loss, geometric scaling
loss, and semantic scaling loss. The overall objective is formulated as
\begin{equation}
    \begin{aligned}
    \mathcal{L} ={} \mathcal{L}_{det}
    + \lambda_{depth}\mathcal{L}_{depth}
    + \lambda\mathcal{L}^{S}_{occ}
    + \sum_i w_i\mathcal{L}^{i}_{occ}.
    \end{aligned}
\end{equation}
\(\mathcal{L}_{det}\) supervises the
detection branch,
\(\mathcal{L}^{S}_{occ}\)
supervises the final occupancy output, and the weighted intermediate losses
supervise state predictions at each stage. \(\mathcal{L}_{depth}\) is used when
depth supervision is available. Further implementation details are provided in the
supplementary materials.

\begin{table}[t]
    \belowrulesep=0pt
    \aboverulesep=0pt
    \centering
    \footnotesize
    \renewcommand\arraystretch{1.00}
    \setlength{\tabcolsep}{0pt}
    \begin{tabular}{@{}C{4.10cm}|C{0.85cm}|C{1.08cm}|C{1.08cm}|C{1.08cm}@{}}
    \toprule[1.0pt]
    Method & Mod. & mAP$\uparrow$ & ODS$\uparrow$ & mIoU$\uparrow$ \\
    \midrule
    BEVFormer-S (ECCV 2022) & C & 28.11 & 29.36 & 15.03 \\
    BEVFormer (ECCV 2022) & C & 30.39 & 31.61 & 15.84 \\
    PanoOcc (CVPR 2024) & C & 26.09 & 27.16 & 14.02 \\
    BEVFusion (NeurIPS 2022) & C\&R & 35.83 & 44.95 & 15.36 \\
    M-CONet (ICCV 2023) & C\&R & -- & -- & 15.30 \\
    RCBEVDet (CVPR 2024) & C\&R & 37.49 & \thirdcell{47.32} & -- \\
    Doracamom-S (TCSVT 2026) & C\&R & \thirdcell{38.75} & 43.47 & \thirdcell{18.81} \\
    Doracamom (TCSVT 2026) & C\&R & \secondcell{41.86} & \secondcell{48.74} & \secondcell{20.30} \\
    \cellcolor{gray!20}\method{} \textbf{(ours)} & \cellcolor{gray!20}C\&R & \bestcell{44.18} & \bestcell{51.57} & \bestcell{23.17} \\
    \bottomrule[1.0pt]
    \end{tabular}
    \vspace{-5pt}
    \caption{OmniHD-Scenes adverse-condition robustness.}
    \label{tab_adverse_main}
    \vspace{-15pt}
\end{table}

\vspace{-5pt}
\section{Experiments}

\subsection{Experimental Setup}

\paragraph{Datasets and Metrics.}
We evaluate \method{} on multi-view OmniHD-Scenes and ManTruckScenes. OmniHD-Scenes uses
six cameras and six 4D radars and provides 4-class detection labels with
11-class semantic occupancy. ManTruckScenes provides a complementary
commercial-vehicle setting with four cameras and our generated 10-class
occupied semantic export.
The export is audited over 523 scenes (464 default, 42 optional inspection, 17
excluded) with automatic scoring and manual evidence-bundle review. Detection is
reported with mAP, dataset-specific overall score
(ODS for OmniHD-Scenes and NDS for ManTruckScenes),
mATE, mASE, mAOE, mAVE, per-class AP, and FPS when available. Occupancy is
reported with SC IoU, mIoU, and per-class IoU. We also evaluate night and rainy
OmniHD-Scenes subsets; details are provided in the supplementary material.

\paragraph{Implementation Details.}
All methods share one MMDetection3D contract: synchronized inputs, temporal
metadata, Det/Occ labels, masks, voxel ranges, optimizer schedule,
augmentation, batch size, and evaluation hooks. Missing Det/Occ outputs are
added only through task-minimal heads while keeping each baseline's original
perception path. Training uses AdamW for 16 epochs. Both datasets use the same
point-cloud range and \(160\times240\times16\) grid. All ablations are built on
the adapted RCBEVDet baseline.

\paragraph{Claim-Evidence Alignment.}
The evaluation directly tests four claims: Tables \ref{tab_omnihd_det}--\ref{tab_mantruck_occ_export}
measure whether state reasoning improves both detection and occupancy across
datasets; Table \ref{tab_adverse_main} tests the radar-camera advantage under
rainy and night conditions; Table \ref{tab_ablation_main} isolates the
state/SBE/DTF contribution; and Tables \ref{tab_ablation_horizon_main}--\ref{tab_ablation_doppler_main}
verify that temporal gains come from state-aware memory and Doppler correction
rather than generic history fusion.

\begin{table}[t]
    \centering
    \footnotesize
    \setlength{\tabcolsep}{0pt}
    \belowrulesep=0pt
    \aboverulesep=0pt
    \renewcommand\arraystretch{1.00}
    \begin{tabular}{@{}C{1.25cm}|C{1.35cm}C{1.15cm}C{1.15cm}|C{1.15cm}C{1.15cm}C{1.15cm}@{}}
    \toprule[1.0pt]
    Base & State & SBE & DTF & mAP$\uparrow$ & NDS$\uparrow$ & mIoU$\uparrow$\\
    \midrule
    \cmark & -- & -- & -- & 41.72 & 46.85 & 28.69\\
    \cmark & \cmark & -- & -- & 45.12 & 49.24 & 29.62\\
    \cmark & \cmark & \cmark & -- & 47.45 & 50.89 & 30.22\\
    \cmark & \cmark & \cmark & \cmark & 49.57 & 52.97 & 31.65\\
    \bottomrule[1.0pt]
    \end{tabular}
    \vspace{-5pt}
    \caption{Component ablation.}
    \label{tab_ablation_main}
    \vspace{-10pt}
    \end{table}

\vspace{-5pt}
\subsection{3D Object Detection Results}
Tables \ref{tab_omnihd_det} and \ref{tab_mantruck_det} show
that the detection gain is not limited to one dataset. On OmniHD-Scenes,
\method{} improves over the strongest prior radar-camera row by 3.32 mAP and
3.70 ODS, with the largest class gain on large vehicles. On
ManTruckScenes, the margin is larger: \method{} exceeds HGSFusion by 5.67 mAP
and 4.74 NDS, while improving the obstacle AP by 15.20 over the best baseline.
The error metrics further localize the source of the gain. mATE and mAVE drop to
0.3613 and 0.9442 on ManTruckScenes, indicating better localization and motion
estimation.

\vspace{-5pt}
\subsection{Semantic Occupancy Prediction Results}
Tables \ref{tab_omnihd_occ} and
\ref{tab_mantruck_occ_export} show complementary occupancy behavior. On
OmniHD-Scenes, \method{} reaches best results, improving over
Doracamom by 1.09 SC IoU and 2.90 mIoU. The gains are strongest on thin or
foreground classes, while driveable surface and sidewalk
also improves. On ManTruckScenes, \method{} gives the best SC
IoU and improves mIoU by +2.29 over HGSFusion.

\vspace{-5pt}
\subsection{Adverse Conditions}
Table \ref{tab_adverse_main} evaluates night and rainy
OmniHD-Scenes subsets. \method{} reaches best performance, improving over Doracamom by 2.32, 2.83, and 2.87 points. These
adverse scenes emphasize the regime where 4D radar is complementary to cameras,
since radar returns remain sparse but stable when image appearance is degraded
by illumination or weather.

\vspace{-5pt}
\subsection{Ablation Studies}
Ablations are conducted on ManTruckScenes. Further details
are reported in the supplementary materials. Detailed resource and efficiency comparisons are provided in the supplementary
materials, showing that the accuracy gains do not come from a resource increase.

\begin{table}[t]
\belowrulesep=0pt
\aboverulesep=0pt
\centering
\footnotesize
\renewcommand\arraystretch{1.00}
\setlength{\tabcolsep}{0pt}

\begin{tabular}{@{}C{1.68cm}|C{0.86cm}C{0.86cm}C{0.86cm}|C{1.30cm}C{1.30cm}C{1.30cm}@{}}
    \toprule[1.0pt]
    Variant & Fr. & State & Dop. & mAP$\uparrow$ & NDS$\uparrow$ & mIoU$\uparrow$ \\
    \midrule
    Single & 1 & -- & -- & 47.45 & 50.89 & 30.22 \\
    BEV-TF & 2 & -- & -- & 46.98 & 50.26 & 29.91 \\
    DTF & 2 & \cmark & \cmark & 48.62 & 52.05 & 30.74 \\
    DTF & 4 & \cmark & \cmark & 49.57 & 52.97 & 31.65 \\
    \bottomrule[1.0pt]
\end{tabular}
\vspace{-5pt}
\caption{Temporal horizon ablation.}
\label{tab_ablation_horizon_main}
\vspace{-10pt}
\end{table}

\begin{table}[t]
\belowrulesep=0pt
\aboverulesep=0pt
\centering
\footnotesize
\renewcommand\arraystretch{1.00}
\setlength{\tabcolsep}{0pt}
\begin{tabular}{@{}C{1.62cm}|C{0.62cm}C{0.62cm}C{0.62cm}C{0.62cm}|C{1.0cm}C{1.0cm}C{1.0cm}C{1.0cm}@{}}
    \toprule[1.0pt]
    Variant & Fr. & Ego & State & Dop. & mAP$\uparrow$ & NDS$\uparrow$ & mIoU$\uparrow$ & mAVE$\downarrow$ \\
    \midrule
    No Temp. & 1 & -- & -- & -- & 47.45 & 50.89 & 30.22 & 1.0213 \\
    Ego Align & 4 & \cmark & -- & -- & 47.12 & 50.45 & 29.96 & 0.9986 \\
    State Mem. & 4 & \cmark & \cmark & -- & 49.06 & 52.28 & 30.92 & 0.9824 \\
    DTF & 4 & \cmark & \cmark & \cmark & 49.57 & 52.97 & 31.65 & 0.9442 \\
    \bottomrule[1.0pt]
\end{tabular}
\vspace{-5pt}
\caption{Doppler ablation.}
\label{tab_ablation_doppler_main}
\vspace{-10pt}
\end{table}

\noindent\textbf{Component contribution.}
Table \ref{tab_ablation_main} shows that the first improvement appears as soon as
occupancy is organized as a coarse-to-fine state path. This row adds intermediate
state prediction, stage propagation, and occupancy-aware voxel recovery, bringing
3.40 mAP, 2.39 NDS, and 0.93 mIoU over the baseline. Then, SBE adds another
2.33 mAP, 1.65 NDS, and 0.60 mIoU, confirming that feeding the explicit state
back to BEV matters for both boxes and voxels. DTF further adds 2.12 mAP,
2.08 NDS, and 1.43 mIoU, giving a total improvement of 7.85/6.12/2.96 over the
baseline.

\noindent\textbf{Temporal horizon.}
Table \ref{tab_ablation_horizon_main} shows that simply adding history is
harmful: two-frame BEV-TF drops below the single-frame model by 0.47 mAP and
0.31 mIoU. DTF reverses this trend because it selects and aligns history through
the learned state. Two-frame DTF improves over single-frame by 1.17 mAP and
0.52 mIoU, while four-frame DTF gives the best result, showing that temporal
context is useful only when history is retrieved with state and motion guidance.

\noindent\textbf{Doppler effectiveness.}
Table \ref{tab_ablation_doppler_main} makes the motion effect explicit. Ego
alignment alone reduces mAVE from 1.0213 to 0.9986 but hurts mAP, NDS, and
mIoU, showing that ego compensation cannot handle moving objects by itself.
Explicit state memory recovers the metrics, and then Doppler gives the decisive
motion correction: mAP, NDS, and mIoU further increase by 0.51, 0.69, and
0.73, while mAVE drops from 0.9824 to 0.9442.

\section{Conclusion}
In this work, we presented \method{}, a systematic 4D radar-camera framework for 360$^\circ$
multi-view multi-task perception. Its main contribution is occupancy-state
reasoning, which models semantic occupancy as a persistent scene state rather
than a terminal output for radar-camera representation learning. SBE and DTF
make this state useful within and across frames, allowing foreground evidence,
dense layout, and radar motion to reinforce a common full-scene representation.
This extends occupancy beyond an auxiliary prediction toward an intermediate reasoning interface that
couples object detection with comprehensive scene understanding. Experiments on OmniHD-Scenes
and ManTruckScenes show consistent gains in detection and occupancy prediction,
while the latter occupancy export provides a second benchmark contract
for future 4D radar-camera multi-task studies. 

\emph{Limitation.} The current work is still a
perception-stage framework and does not yet train an end-to-end planning or VLA
system; extending the learned radar-aware state to those settings is the next
step.

\bibliography{aaai2027}

@ARTICLE{SGDet3D,
  author={Bai, Xiaokai and Yu, Zhu and Zheng, Lianqing and Zhang, Xiaohan and Zhou, Zili and Zhang, Xue and Wang, Fang and Bai, Jie and Shen, Hui-Liang},
  journal={IEEE Robotics and Automation Letters}, 
  title={{SGDet3D: Semantics and Geometry Fusion for 3D Object Detection Using 4D Radar and Camera}}, 
  year={2024},
  volume={},
  number={},
  pages={1-8},
  doi={10.1109/LRA.2024.3513041}}

@inproceedings{RaGS2026,
  title={{RaGS: Unleashing 3D Gaussian Splatting from 4D Radar and Monocular Cue for 3D Object Detection}},
  author={Bai, Xiaokai and Zhou, Chenxu and Zheng, Lianqing and Cao, Si-Yuan and Liu, Jianan and Zhang, Xiaohan and Li, Yiming and Zhang, Zhengzhuang and Shen, Hui-Liang},
  booktitle={IEEE/CVF Conference on Computer Vision and Pattern Recognition},
  year={2026}
}

@article{R4Det2026,
  title={{R4Det: 4D Radar-Camera Fusion for High-Performance 3D Object Detection}},
  author={Xia, Zhongyu and Tang, Yousen and Wang, Yongtao and Wang, Zhifeng and Qin, Weijun},
  journal={arXiv preprint arXiv:2603.11566},
  year={2026}
}

@inproceedings{LGDD2025,
  title={{LGDD}: Local-Global Synergistic Dual-Branch 3D Object Detection Using 4D Radar},
  author={Bai, Xiaokai and Yang, Qin and Zhou, Zili and Zhang, Fuyi and Wu, Zhe and Cao, Si-Yuan and Zheng, Lianqing and Yu, Beinan and Wang, Fang and Bai, Jie and Shen, Hui-Liang},
  booktitle={IEEE/RSJ International Conference on Intelligent Robots and Systems},
  pages={13318--13325},
  year={2025}
}

@inproceedings{SD4R2026,
  title={{SD4R}: Sparse-to-Dense Learning for 3D Object Detection with 4D Radar},
  author={Bai, Xiaokai and Cheng, Jiahao and Wang, Songkai and Luo, Yixuan and Zheng, Lianqing and Zhang, Xiaohan and Cao, Si-Yuan and Shen, Hui-Liang},
  booktitle={IEEE International Conference on Intelligent Transportation Systems},
  pages={4362--4368},
  year={2025}
}

@article{WRCFormer2025,
  title={{Wavelet-based Multi-View Fusion of 4D Radar Tensor and Camera for Robust 3D Object Detection}},
  author={Guan, Runwei and Liu, Jianan and Liang, Shaofeng and Ding, Fangqiang and Yao, Shanliang and Bai, Xiaokai and Liu, Daizong and Huang, Tao and Mao, Guoqiang and Xiong, Hui},
  journal={arXiv preprint arXiv:2512.22972},
  year={2025}
}

@inproceedings{SARCD2025,
  title={{Structure-Aware Radar-Camera Depth Estimation}},
  author={Zhang, Fuyi and Yu, Zhu and Li, Chunhao and Zhang, Runmin and Bai, Xiaokai and Zhou, Zili and Cao, Si-Yuan and Wang, Fang and Shen, Hui-Liang},
  booktitle={IEEE International Conference on Robotics and Automation},
  pages={13028--13035},
  year={2025}
}

@article{SIFormer2026,
  title={{Boosting Instance Awareness via Cross-View Correlation with 4D Radar and Camera for 3D Object Detection}},
  author={Bai, Xiaokai and Zheng, Lianqing and Cao, Si-Yuan and Zhang, Xiaohan and Wu, Zhe and Yu, Beinan and Wang, Fang and Bai, Jie and Shen, Hui-Liang},
  journal={arXiv preprint arXiv:2602.20632},
  year={2026}
}

@article{RCGeoCP2026,
  title={{RC-GeoCP}: Geometric Consensus for Radar-Camera Collaborative Perception},
  author={Bai, Xiaokai and Zheng, Lianqing and Guan, Runwei and Cao, Si-Yuan and Shen, Hui-Liang},
  journal={arXiv preprint arXiv:2603.00654},
  year={2026}
}

@article{RCFusion,
  title={{RCFusion: Fusing 4-D Radar and Camera with Bird’s-Eye View Features for 3-D Object Detection}},
  author={Zheng, Lianqing and Li, Sen and Tan, Bin and Yang, Long and Chen, Sihan and Huang, Libo and Bai, Jie and Zhu, Xichan and Ma, Zhixiong},
  journal={IEEE Transactions on Instrumentation and Measurement},
  volume={72},
  pages={1--14},
  year={2023},
  publisher={IEEE}
}

@article{SurroundOcc,
  title={{SurroundOcc: Multi-Camera 3D Occupancy Prediction for Autonomous Driving}},
  author={Wei, Yi and Zhao, Linqing and Zheng, Wenzhao and Zhu, Zheng and Zhou, Jie and Lu, Jiwen},
  journal={arXiv preprint arXiv:2303.09551},
  year={2023}
}

@article{OmniHD,
  title={{OmniHD-Scenes: A Next-Generation Multimodal Dataset for Autonomous Driving}},
  author={Zheng, Lianqing and Yang, Long and Lin, Qunshu and Ai, Wenjin and Liu, Minghao and Lu, Shouyi and Liu, Jianan and Ren, Hongze and Mo, Jingyue and Bai, Xiaokai and others},
  journal={arXiv preprint arXiv:2412.10734},
  year={2024}
}

@inproceedings{BEVFusion,
  title={{BEVFusion: Multi-Task Multi-Sensor Fusion with Unified Bird’s-Eye View Representation}},
  author={Liu, Zhijian and Tang, Haotian and Amini, Alexander and Yang, Xinyu and Mao, Huizi and Rus, Daniela L and Han, Song},
  booktitle={IEEE International Conference on Robotics and Automation},
  pages={2774--2781},
  year={2023}
}

@article{LXL,
  author={Xiong, Weiyi and Liu, Jianan and Huang, Tao and Han, Qing-Long and Xia, Yuxuan and Zhu, Bing},
  journal={IEEE Transactions on Intelligent Vehicles}, 
  title={{LXL: LiDAR Excluded Lean 3D Object Detection with 4D Imaging Radar and Camera Fusion}}, 
  year={2024},
  volume={9},
  number={1},
  pages={79-92},
  doi={10.1109/TIV.2023.3321240}}

@inproceedings{HGSFusion,
  title={{HGSFusion: Radar-Camera Fusion with Hybrid Generation and Synchronization for 3D Object Detection}},
  author={Gu, Zijian and Ma, Jianwei and Huang, Yan and Wei, Honghao and Chen, Zhanye and Zhang, Hui and Hong, Wei},
  booktitle={AAAI Conference on Artificial Intelligence},
  volume={39},
  pages={3185--3193},
  year={2025}
}

@inproceedings{centerfusion,
  title={{CenterFusion: Center-based Radar and Camera Fusion for 3D Object Detection}},
  author={Nabati, Ramin and Qi, Hairong},
  booktitle={IEEE/CVF Winter Conference on Applications of Computer Vision},
  pages={1527--1536},
  year={2021}
}

@inproceedings{CRAFT,
  title={{CRAFT: Camera-Radar 3D Object Detection with Spatio-Contextual Fusion Transformer}},
  author={Kim, Youngseok and Kim, Sanmin and Choi, Jun Won and Kum, Dongsuk},
  booktitle={AAAI Conference on Artificial Intelligence},
  volume={37},
  pages={1160--1168},
  year={2023}
}

@article{VoD,
  title={{Multi-Class Road User Detection with 3+1D Radar in the View-of-Delft Dataset}},
  author={Palffy, Andras and Pool, Ewoud and Baratam, Srimannarayana and Kooij, Julian FP and Gavrila, Dariu M},
  journal={IEEE Robotics and Automation Letters},
  volume={7},
  number={2},
  pages={4961--4968},
  year={2022},
  publisher={IEEE}
}

@inproceedings{TJ4D,
  title={{TJ4DRadSet: A 4D Radar Dataset for Autonomous Driving}},
  author={Zheng, Lianqing and Ma, Zhixiong and Zhu, Xichan and Tan, Bin and Li, Sen and Long, Kai and Sun, Weiqi and Chen, Sihan and Zhang, Lu and Wan, Mengyue and others},
  booktitle={IEEE International Conference on Intelligent Transportation Systems},
  pages={493--498},
  year={2022}
}

@inproceedings{RCBEVDet,
  title={{RCBEVDet: Radar-camera Fusion in Bird's Eye View for 3D Object Detection}},
  author={Lin, Zhiwei and Liu, Zhe and Xia, Zhongyu and Wang, Xinhao and Wang, Yongtao and Qi, Shengxiang and Dong, Yang and Dong, Nan and Zhang, Le and Zhu, Ce},
  booktitle={IEEE/CVF Conference on Computer Vision and Pattern Recognition},
  pages={14928--14937},
  year={2024}
}

@article{SECOND,
  title={{SECOND: Sparsely Embedded Convolutional Detection}},
  author={Yan, Yan and Mao, Yuxing and Li, Bo},
  journal={Sensors},
  volume={18},
  number={10},
  pages={3337},
  year={2018},
  publisher={Multidisciplinary Digital Publishing Institute}
}

@inproceedings{monoscene,
  title={{MonoScene: Monocular 3D Semantic Scene Completion}},
  author={Cao, Anh-Quan and De Charette, Raoul},
  booktitle={IEEE/CVF Conference on Computer Vision and Pattern Recognition},
  pages={3991--4001},
  year={2022}
}

@inproceedings{voxformer,
  title={VoxFormer: Sparse Voxel Transformer for Camera-based 3D Semantic Scene Completion},
  author={Li, Yiming and Yu, Zhiding and Choy, Christopher and Xiao, Chaowei and Alvarez, Jose M and Fidler, Sanja and Feng, Chen and Anandkumar, Anima},
  booktitle={IEEE/CVF Conference on Computer Vision and Pattern Recognition},
  pages={9087--9098},
  year={2023}
}

@inproceedings{RADIANT,
  title={{RADIANT: Radar-Image Association Network for 3D Object Detection}},
  author={Long, Yunfei and Kumar, Abhinav and Morris, Daniel and Liu, Xiaoming and Castro, Marcos and Chakravarty, Punarjay},
  booktitle={AAAI Conference on Artificial Intelligence},
  volume={37},
  pages={1808--1816},
  year={2023}
}

@article{Kradar,
  title={{K-Radar: 4D Radar Object Detection for Autonomous Driving in Various Weather Conditions}},
  author={Paek, Dong-Hee and Kong, Seung-Hyun and Wijaya, Kevin Tirta},
  journal={Advances in Neural Information Processing Systems},
  volume={35},
  pages={3819--3829},
  year={2022}
}

@inproceedings{Gaussianformer,
  title={{GaussianFormer: Scene as Gaussians for Vision-Based 3D Semantic Occupancy Prediction}},
  author={Huang, Yuanhui and Zheng, Wenzhao and Zhang, Yunpeng and Zhou, Jie and Lu, Jiwen},
  booktitle={European Conference on Computer Vision},
  pages={376--393},
  year={2024},
  organization={Springer}
}

@inproceedings{wang2024panoocc,
  title={{PanoOcc}: Unified Occupancy Representation for Camera-Based 3D Panoptic Segmentation},
  author={Wang, Yuqi and Chen, Yuntao and Liao, Xingyu and Fan, Lue and Zhang, Zhaoxiang},
  booktitle={IEEE/CVF Conference on Computer Vision and Pattern Recognition},
  pages={17158--17168},
  year={2024}
}

@inproceedings{wang2023openoccupancy,
  title={{OpenOccupancy}: A Large Scale Benchmark for Surrounding Semantic Occupancy Perception},
  author={Wang, Xiaofeng and Zhu, Zheng and Xu, Wenbo and Zhang, Yunpeng and Wei, Yi and Chi, Xu and Ye, Yun and Du, Dalong and Lu, Jiwen and Wang, Xingang},
  booktitle={IEEE/CVF International Conference on Computer Vision},
  pages={17850--17859},
  year={2023}
}

@article{Doracamom2026,
  title={{Doracamom}: Joint 3D Detection and Occupancy Prediction with Multi-view 4D Radars and Cameras for Omnidirectional Perception},
  author={Zheng, Lianqing and Liu, Jianan and Guan, Runwei and Yang, Long and Lu, Shouyi and Li, Yuanzhe and Bai, Xiaokai and Bai, Jie and Ma, Zhixiong and Shen, Hui-Liang and Zhu, Xichan},
  journal={IEEE Transactions on Circuits and Systems for Video Technology},
  year={2026}
}

@inproceedings{SOGDet2024,
  title={{SOGDet}: Semantic-Occupancy Guided Multi-view 3D Object Detection},
  author={Zhou, Qiu and Cao, Jinming and Leng, Hanchao and Yin, Yifang and Kun, Yu and Zimmermann, Roger},
  booktitle={AAAI Conference on Artificial Intelligence},
  year={2024}
}

@article{UniVision2024,
  title={{UniVision}: A Unified Framework for Vision-Centric 3D Perception},
  author={Hong, Yu and Liu, Qian and Cheng, Huayuan and Ma, Danjiao and Dai, Hang and Wang, Yu and Cao, Guangzhi and Ding, Yong},
  journal={arXiv preprint arXiv:2401.06994},
  year={2024}
}

@article{Occ3D2023,
  title={{Occ3D}: A Large-Scale 3D Occupancy Prediction Benchmark for Autonomous Driving},
  author={Tian, Xiaoyu and Jiang, Tao and Yun, Longfei and Mao, Yucheng and Yang, Huitong and Wang, Yue and Wang, Yilun and Zhao, Hang},
  journal={Advances in Neural Information Processing Systems},
  year={2023}
}
\end{document}